\documentclass{article}

\usepackage{microtype}
\usepackage{graphicx}
\usepackage{subfigure}
\usepackage{booktabs} 
\usepackage[most]{tcolorbox} 
\tcbset{
  mytemplatebox/.style={
    colback=gray!5,
    colframe=black,
    fonttitle=\bfseries,
    coltitle=black,
    sharp corners,
    boxrule=0.4pt,
    enhanced,
    breakable,
    listing only,
    listing options={
      basicstyle=\ttfamily\small,
      breaklines=true,
      breakatwhitespace=true,
      columns=fullflexible,
      frame=none,
      escapeinside={(*@}{@*)}
    },
    before skip=10pt,
    after skip=10pt,
  }
}
\usepackage{enumitem}
\usepackage{hyperref}

\usepackage[accepted]{icml2025}
\makeatletter
\renewcommand{\ICML@appearing}{Accepted at the ICML 2025 Workshop on Multi-Agent Systems in the Era of Foundation Models: Opportunities, Challenges and Futures (MAS-2025).}
\makeatother

\usepackage{amsmath}
\usepackage{amssymb}
\usepackage{mathtools}
\usepackage{amsthm}

\usepackage[capitalize,noabbrev]{cleveref}

\theoremstyle{plain}

\theoremstyle{definition}

\theoremstyle{remark}

\usepackage[textsize=tiny]{todonotes}

\icmltitlerunning{MIRAGE: Multi-Perspective Creative Language Model Reasoning}

\begin{document}

\twocolumn[
\icmltitle{MIRAGE:  Multi-Perspective Creative Language Model Reasoning with Reinforcement Learning Guidance}




\begin{icmlauthorlist}
\icmlauthor{Arash Lagzian}{nus}
\icmlauthor{Srinivas Anumasa}{nus}
\icmlauthor{Dianbo Liu}{nus}
\end{icmlauthorlist}

\icmlaffiliation{nus}{National University of Singapore}

\icmlcorrespondingauthor{Arash Lagzian}{alagzian@visitor.nus.edu.sg}
\icmlcorrespondingauthor{Srinivas Anumasa}{srinu\_pd@nus.edu.sg}
\icmlcorrespondingauthor{Dianbo Liu}{dianbo@nus.edu.sg}
\icmlkeywords{Machine Learning, ICML}

\vskip 0.3in
]



\printAffiliationsAndNotice{}

\begin{abstract}
Recent advances in Large Language Models (LLMs) have revolutionized artificial intelligence and how human interact with AIs.  Despite impressive advancements, LLMs struggle with complex mathematical, scientific, and logical tasks. Inspired by human cognitive flexibility—our ability to dynamically switch mental perspectives—we propose \textbf{MIRAGE} (\textbf{M}ulti-perspective \textbf{I}nference-time \textbf{R}easoning via \textbf{A}gent-\textbf{G}uided \textbf{E}xploration), a novel inference-time creative thinking framework. MIRAGE includes a \textit{Selector} that prioritizes effective conceptual perspectives (e.g., algebraic, probabilistic) and a \textit{Reasoner} that sequentially solves tasks until a confident solution emerges, otherwise aggregating multiple perspectives. Tested on GSM8K, MATH500, MMLU-Pro, and Game-of-24 benchmarks, MIRAGE consistently outperforms methods like Chain-of-Thought and diverse prompting ensembles, significantly boosting accuracy with minimal inference overhead, providing a scalable solution for practical applications.
\end{abstract}

\section{Introduction}
Recent progress in Large Language Models (LLMs) has greatly improved their ability to handle open-ended language tasks, but they still struggle with complex reasoning in math, science, and logic. Even small changes in wording or how the model generates its response can cause it to go from a correct answer to a completely wrong one. This kind of fragility makes it hard to trust LLMs in important applications like tutoring systems, engineering assistants, or financial tools \citep{cobbe2021training, wei2022chain}. To close this gap, researchers have explored ever richer prompting and inference-time techniques: Chain-of-Thought (CoT) prompting steers models through explicit reasoning steps \citep{wei2022chain}, zero-shot CoT and scratchpads remove the need for demonstrations \citep{kojima2022large, nye2022scratchpads}, and Least-to-Most prompting decomposes complex problems into simpler sub-questions \citep{zhou2023least}. Decoding strategies such as Self-Consistency \citep{wang2023selfconsistency}, ensemble methods like Reflexion and DIPPER sample multiple reasoning paths and vote on an answer \citep{shinn2023reflexion, lau2024dipper}, while tool-augmented frameworks call external calculators or verifiers to patch errors \citep{schick2023toolformer}.

\textbf{Why do existing fixes fall short?} \emph{Fragile prompting}—CoT variants depend on carefully crafted exemplars; minor edits can derail generation \citep{kojima2022large}. \emph{Costly ensembling}—sampling three to five reasoning paths per query boosts accuracy but inflates latency and API cost by up to \(5\times\) \citep{wang2023selfconsistency, lau2024dipper}. \emph{Rigid single-perspective reasoning}—all methods process the problem through one fixed representation; if that perspective misaligned with the task’s structure, there is no fallback.  Tool-augmented or fine-tuned systems further add infrastructure overhead and sacrifice model-agnostic portability.

Decades of research in cognitive science and neuroscience reveal that humans seldom resolve such challenges using a single representational approach. Instead, we fluidly \emph{re-encode} problems—transforming algebraic tasks into geometric ones, or probability tasks into frequency-based representations—until one representation clearly facilitates insight. This principle of cognitive flexibility is supported by studies on strategy switching \cite{siegler1996emerging}, representational shifts underlying insights \cite{knoblich1999constraint}, and evidence for specialized parallel neural circuits \cite{dehaene2009origins, deen2021parallel}.

Addressing the reasoning challenge using inspiration from cognitive science, we introduce \textbf{MIRAGE}, an creative multi-perspective inference-time reasoning framework inspired by the human cognitive strategy of problem re-representation to discover insightful solutions. For mathematic reasoning tasks, MIRAGE employs a \emph{Selector} that ranks twenty conceptual reasoning perspectives—including algebraic, probabilistic, and game-theoretic frameworks—based on their historical effectiveness on similar problems. Subsequently, a \emph{Reasoner} iteratively attempts these perspectives in ranked order, halting either upon reaching a high-confidence solution or after forming a small ensemble. Remarkably, MIRAGE requires no parameter updates to the base LLM and averages fewer than two forward passes on three of four benchmarks, achieving superior accuracy–cost trade-offs (Figure~\ref{fig:accuracy-vs-cost}).

Our contributions in this study are as follows:

\begin{itemize}
    \item We introduce the first inference-time framework that \emph{learns} to choose among conceptual reasoning perspectives (\textbf{Selector}) and solves within them (\textbf{Reasoner}) without modifying LLM weights.  
    \item Across GSM8K, MATH500, MMLU-Pro, and Game-of-24, MIRAGE\ lifts accuracy by up to \textbf{+24.7\,pp} over CoT while using at most \textbf{$2\times$} the cost of a single call and up to \textbf{$5\times$} less cost than DIPPER ($n{=}5$).  
    \item Extensive experiments on five base models—DeepSeek-v3, ChatGPT-4o, Claude 3.7-Sonnet, Gemini-Flash 2.0, and Qwen2.5-7B—confirm consistent gains and detailed accuracy-vs-cost analysis
\end{itemize}

\begin{figure}[!t]
  \centering
  \includegraphics[width=\linewidth]{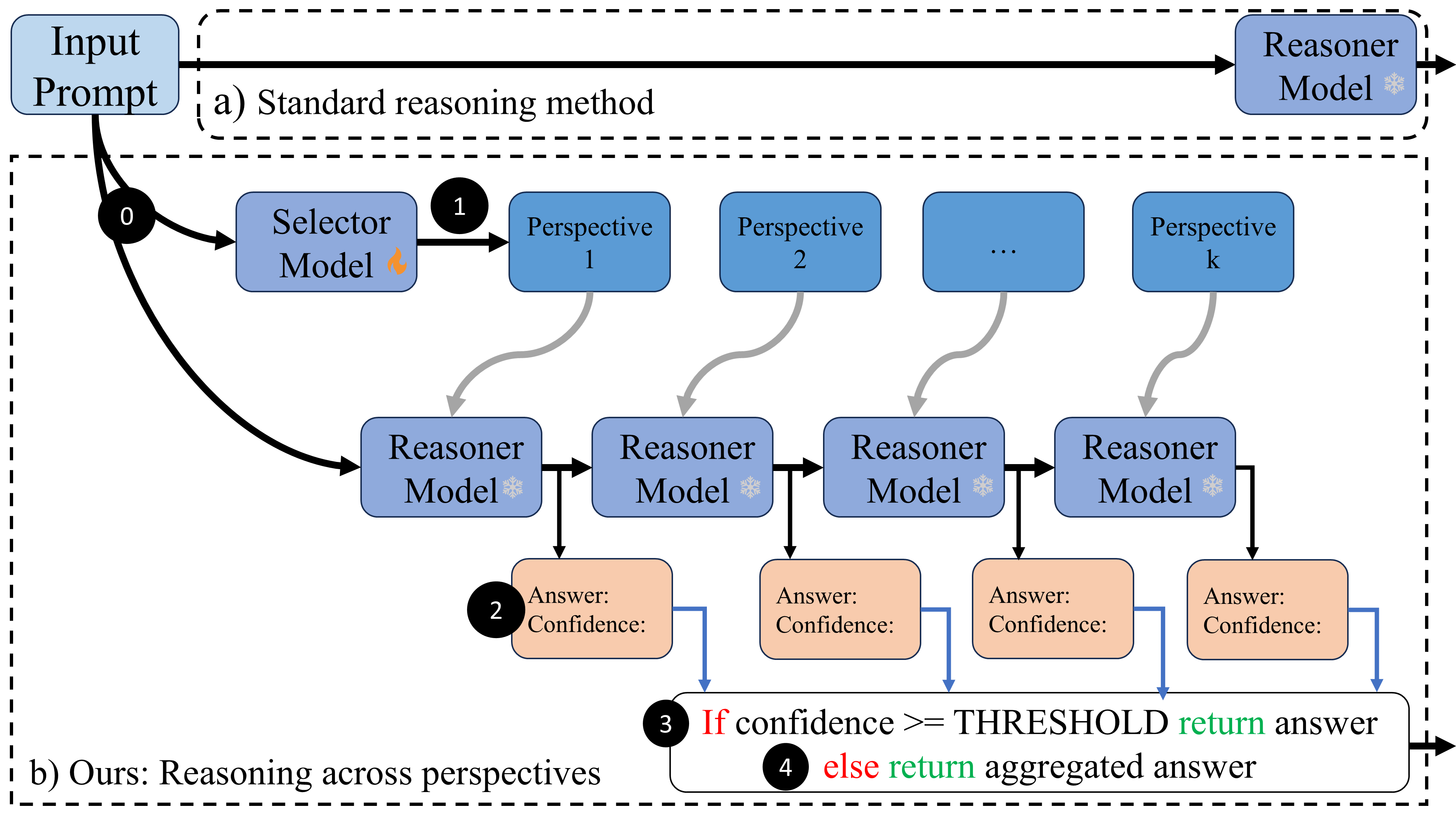}
  \caption{Overview of our multi-perspective reasoning framework.
    \textbf{(a)} Standard reasoning uses a single forward pass through the Reasoner model.
    \textbf{(b)} Our method proceeds in four stages: (1) the Selector ranks conceptual perspectives, (2) the Reasoner model solves each perspective, (3) if confidence exceeds the threshold, the answer is returned, and (4) otherwise, answers are aggregated. These stages are illustrated with numbered circles.}

  \label{fig:overal}
\end{figure}

\section{Related Works}
\label{sec:related}

Reasoning with large language models has advanced along three intertwined threads that culminate in the ideas we pursue in \textsc{MIRAGE}.  First, prompt–engineering techniques expose latent chain-of-thought abilities.  Chain-of-Thought (CoT) prompting~\cite{wei2022chain} and its zero-shot variant “Let’s think step by step”~\cite{kojima2022large} showed that supplying intermediate steps can dramatically lift arithmetic and logical accuracy; later extensions introduced \emph{scratchpads} to reveal hidden computations~\cite{nye2022show}, Least-to-Most decomposition for hierarchical problem solving~\cite{zhou2023least}, and self-training with generated rationales in STaR~\cite{zelikman2023star}.  Second, search-based and self-verification methods improve reliability by exploring multiple reasoning paths: Self-Consistency aggregates sampled solutions~\cite{wang2023selfconsistency}, verifier models prune incorrect chains~\cite{cobbe2021training}, and reflection loops iteratively repair errors~\cite{shinn2023reflexion}.  Third, modular agent frameworks route questions to external tools or expert policies.  Tree-of-Thoughts performs deliberative tree search~\cite{yao2023tree}, ReAct interleaves reasoning with tool calls~\cite{yao2023react}, PAL executes generated code to obtain ground-truth signals~\cite{gao2022pal}, Toolformer learns when to query APIs~\cite{schick2023toolformer}, HuggingGPT orchestrates specialist models~\cite{shen2023hugginggpt}, and HDFlow adaptively chooses between fast and slow solvers~\cite{yao2024hdflow}.  Very recent work pushes routing one step further:  DIPPER emphasis on the importance of diversity in input prompts and ensemble models~\cite{lau2024dipper}, and Atomic Reasoner try to extract atomic facts and reason based on them~\cite{zhang2025atomic}.  

\textsc{MIRAGE} builds on these insights but occupies a distinct niche.  Instead of sampling many chains or invoking external APIs, we maintain a \emph{human-interpretable library of twenty conceptual perspectives} (algebraic, probabilistic, network-flow, \emph{etc.}) and train a lightweight selector—once, on solved examples—to choose the most promising perspective at inference time.  This yields a single-call, multi-perspective solution that matches or surpasses the accuracy–cost Pareto front of tree search, verifier, and modular-tool baselines while requiring no additional model fine-tuning.

\section{Motivation from Brain and Cognitive Science}
\label{sec:motivation}

Our approach is fundamentally motivated by well-established cognitive and neuroscientific principles highlighting the importance of cognitive flexibility and multi-perspective reasoning in effective human problem-solving. Cognitive science research consistently demonstrates that humans excel in complex problem-solving scenarios by dynamically shifting among multiple mental frameworks or representational strategies, adapting flexibly as task demands evolve~\cite{spiro1988cognitive,miyake2000unity,siegler1996emerging}. Translating these insights into AI, we posit that enhancing the reasoning capabilities of Large Language Models (LLMs) similarly requires an inference strategy capable of fluidly switching among diverse conceptual representations or reasoning perspectives.

\textbf{Cognitive Flexibility Theory}, as articulated by Spiro et al.~\cite{spiro1988cognitive}, underscores the necessity of constructing knowledge through multiple, overlapping representations to achieve mastery in complex and ill-structured domains. This theory asserts that learners who regularly restructure knowledge across different conceptual perspectives not only deepen their understanding but also enhance their ability to transfer insights effectively to novel contexts. Parallel evidence from developmental psychology, notably the \textbf{Overlapping Waves Theory} introduced by Siegler~\cite{siegler1996emerging}, further supports this principle. Siegler demonstrated that human learners naturally employ and switch among multiple strategies to solve problems, progressively refining strategy selection through experience. Such flexible use of diverse approaches facilitates robust and generalized problem-solving capabilities.

Empirical and neuroscientific evidence converges on the same lesson: switching representations boosts performance.  In classrooms, students who compare multiple algebraic methods achieve deeper procedural and conceptual mastery than peers taught a single approach~\cite{rittle2007does}; likewise, Bayesian problems become far easier when reframed from probabilities to natural frequencies~\cite{gigerenzer1995improve}.  fMRI studies echo this flexibility, revealing parallel circuits dedicated to social versus spatial reasoning~\cite{deen2021parallel} and a triple-code network for numerical quantity, visual, and verbal processing~\cite{dehaene2009origins}, underscoring the brain’s propensity to recruit whichever representation best fits the task.

Research on insight and social cognition paints a similar picture.  Breakthroughs in “aha!” problems often hinge on abandoning an unproductive framing and relaxing prior constraints~\cite{knoblich1999constraint}, while exposure to multiple cultural contexts broadens mental representations and boosts creative problem-solving~\cite{maddux2009cultural}.  Taken together, these strands suggest that intelligent systems should likewise pivot between representations to overcome impasses and generalize.  \textsc{MIRAGE} operationalises this principle by coupling a Selector that chooses among twenty conceptual perspectives with a Reasoner that solves the problem inside each chosen view, aiming to confer human-like flexibility on large language models.

\section{Proposed Method}
\label{sec:method}

In this section, we present our inference-time framework designed to improve the reasoning abilities of LLMs. The core insight is that complex reasoning problems can be solved more effectively when dynamically transformed into multiple different conceptual perspectives specialized for different reasoning paradigms. 

\subsection{Overall Framework}

Given a reasoning task with input prompt \(q\), our goal is to select an appropriate sequence of conceptual reasoning perspectives and iteratively solve within them until a confident solution is found. Let \(\mathcal{S} = \{ s_1, s_2, \dots, s_{20} \}\) denote the set of predefined conceptual perspectives. These 20 perspectives were carefully curated by domain experts through an in-depth analysis of problem-solving strategies commonly observed in mathematics, science, engineering, and logic. The selection process drew from techniques emphasized in educational curricula and employed by experts across disciplines. We prioritized perspectives that are both cognitively distinctive and broadly applicable. A complete description of all 20 perspectives, including detailed justifications for their inclusion and representative problem examples, is provided in Appendix~\ref{appendix:multi-space}.

Our approach consists of two phases: (1) a \emph{Training Phase}, where the \emph{Selector} is trained to choose relevant reasoning perspectives for each input problem while the Reasoner remains fixed, and (2) an \emph{Inference Phase}, where the trained Selector dynamically guides the Reasoner through selected perspectives.
\subsection{Training Phase: Selector Reinforcement Learning}

\paragraph{Setup.}
We train the selector with REINFORCE on \textbf{5700} MMLU\cite{hendrycks2021measuring} questions
(one hundred samples per subject, shuffled).
The selector is a \texttt{Qwen2.5-7B-Instruct}\cite{qwen25} classifier
fine-tuned with LoRA ($r{=}16$, $\alpha{=}32$).
A frozen \texttt{Qwen2.5-14B-Instruct}\cite{qwen2.5-14b-instruct} reasoner, queried once per selected
perspective, generates the step-by-step solution.
Training uses mini-batches of eight questions
and Adam ($\eta=5\!\times\!10^{-6}$) on a pair of A100 GPUs.

\paragraph{Learning dynamics.}
We train the selector using the REINFORCE algorithm with a fixed Reasoner (Algorithm~\ref{alg:selector_train}). At each step, the selector outputs probabilities over the 20 conceptual perspectives, and a binary mask is sampled to decide which perspectives are activated. The Reasoner attempts the problem under each selected perspective, and a majority vote produces the final answer. The reward signal encourages both correctness and sparsity:
\begin{equation}
r = \mathbb{I}[\hat{a} = a^*] - \lambda \cdot \frac{k}{|\mathcal{S}|}
\label{eq:reward}
\end{equation}

where \(k\) is the number of selected perspectives and \(\lambda = 0.05\) is the penalty coefficient.

\begin{figure}[t]
  \centering
  \includegraphics[width=\linewidth]{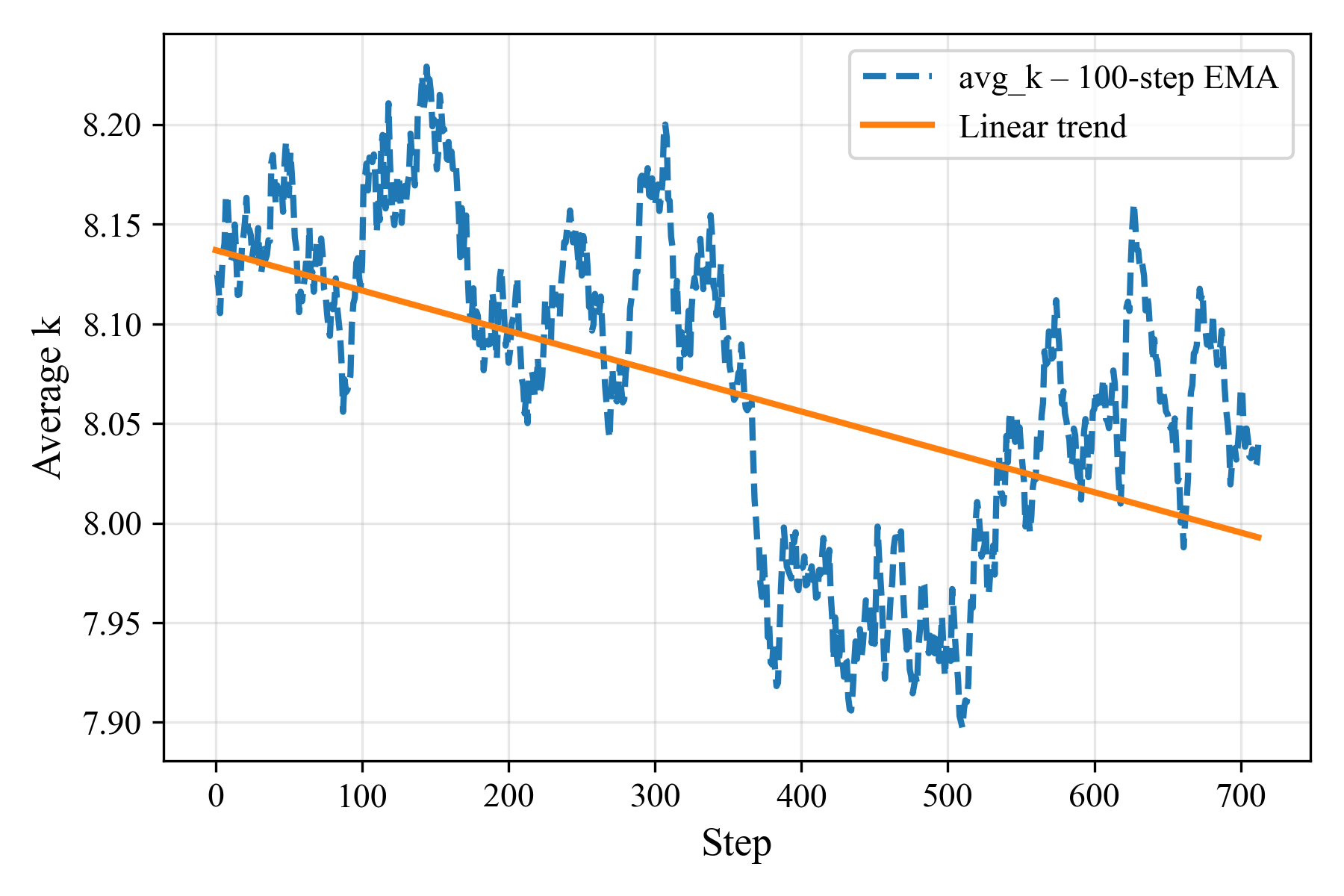}
  \caption{Training curve of the average number of selected perspectives \(k\) during REINFORCE training. The dashed blue line shows the 100‐step EMA of \(k\); the solid orange line is the best‐fit linear trend (slope \(\approx -2.02\times10^{-4}\)), highlighting a slight but consistent downward drift. Under penalty \(\lambda=0.05\), the selector stabilizes at around eight perspectives per query.}
  \label{fig:training_curves}
\end{figure}

Fig.~\ref{fig:training_curves} shows how the average number of perspectives \(k\) selected evolves over training. The blue dashed line tracks the 100-step exponential moving average (EMA) of \(k\), while the orange line shows the best-fit linear trend. Despite fluctuations due to stochastic sampling, the selector exhibits a consistent downward drift in \(k\), stabilizing at around 8 perspectives per query. This emergent sparsity demonstrates the selector’s ability to learn compact yet effective subspaces of reasoning, achieving a 2.5× reduction in reasoning cost compared to querying all 20 perspectives.

\paragraph{Baselines.}
Table~\ref{tab:selector_compare} contrasts our policy
against two baselines on a held-out MMLU slice. Our policy matches the Full-20 method accuracy
while cutting inference cost by 60\%.
It also outperforms a Random-8 selector by
\textbf{+10.9pp}, validating that the model learns
\emph{which} conceptual perspectives matter.

\begin{table}[h]
  \centering
  \caption{Held-out comparison of selector policies.}
  \label{tab:selector_compare}
  \begin{tabular}{lccc}
    \toprule
    \bf Selector & Avg.\,$k$ & Accuracy (\%) & Inference cost \\
    \midrule
    Random-8           & 8   & 63.1 & $1.0\times$ \\
    Full-20            & 20  & 74.5 & $2.5\times$ \\
    \textbf{RL (ours)} & 8.1 & \textbf{74.0} & $1.0\times$ \\
    \bottomrule
  \end{tabular}
\end{table}

\begin{figure}[h]
  \centering
  \includegraphics[width=\linewidth]{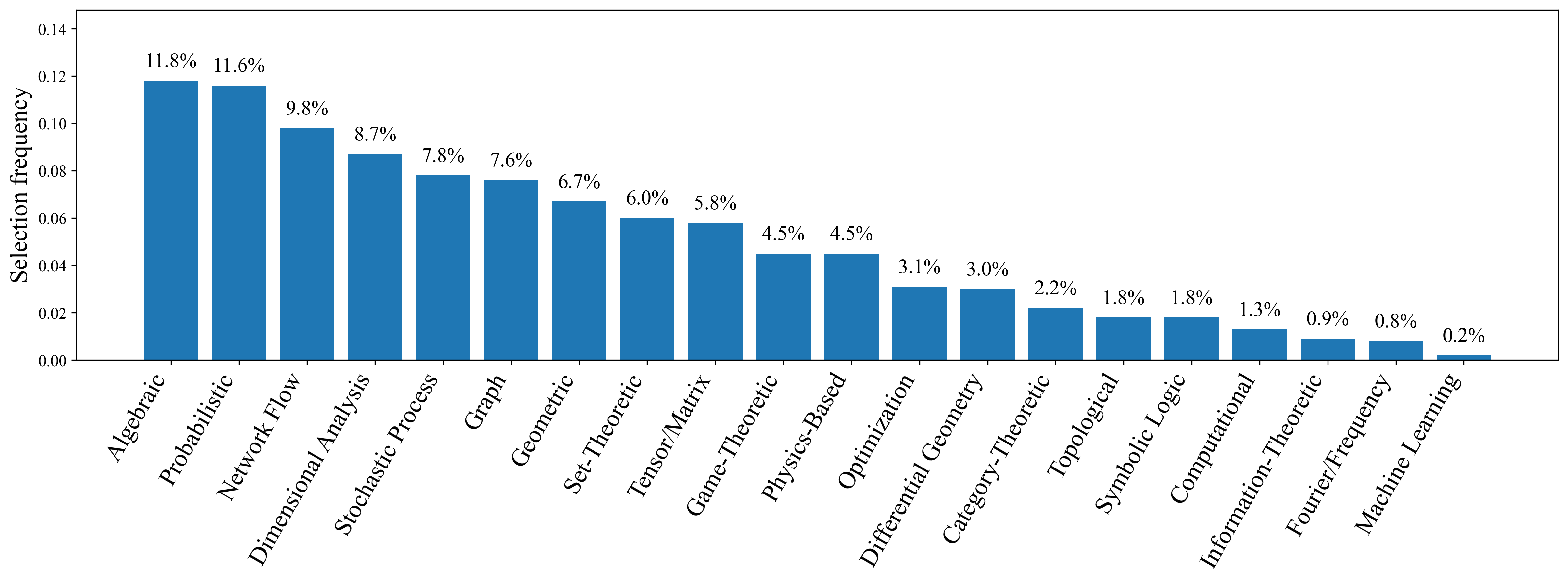}
  \vspace{-6pt}
  \caption{\textbf{Final perspective-Selection Distribution.}
           Normalized usage rates of each conceptual perspective after REINFORCE training ( $\lambda=0.05$). The height of each bar is the fraction of queries the selector directed to that perspective; numerical labels show percentage frequency. Note the clear peak at Algebraic and Probabilistic perspective, followed by a long-tail distribution across the remaining 18 perspective.}
  \label{fig:space_dist}
\end{figure}

\paragraph{Perspective preferences.}
As shown in Fig.~\ref{fig:space_dist}, the trained selector develops a strong preference for a subset of highly predictive reasoning perspectives. The top five—\emph{Algebraic} (11.8\%), \emph{Probabilistic} (11.6\%), \emph{Network Flow} (9.8\%), \emph{Dimensional Analysis} (8.7\%), and \emph{Stochastic Process} (7.8\%)—account for nearly half of all selections. This skewed distribution highlights the effectiveness of the learned policy in focusing computation on a few highly informative perspectives while avoiding low-utility ones. The long tail across the remaining perspectives suggests retained flexibility, allowing the system to fall back on niche reasoning modes when needed.

\subsection{Inference Phase: Dynamic perspective Selection and Sequential Reasoning}
\begin{algorithm}[tb]
\caption{Inference-Time Multi-Perspective Reasoning}
\label{alg:multi-perspective}
\begin{algorithmic}[1]
\REQUIRE Query $q$, perspective set $\mathcal{S}$, selector $\textsc{Selector}$, reasoner $\textsc{Reasoner}$, confidence threshold $\tau$, max attempts $k$
\STATE \textbf{(0)} Receive query $q$
\STATE \textbf{(1)} $S_{\text{ranked}} \gets \textsc{Selector}(q)$ \hfill \COMMENT{Rank perspectives by $p(s_i \mid q)$}
\STATE Initialize history $H \gets \emptyset$
\FOR{$t = 1$ \text{TO} $k$}
    \STATE $s_{(t)} \gets S_{\text{ranked}}[t]$ \hfill \COMMENT{Select top-ranked perspective}
    \STATE $q_{(t)} \gets \textsc{Transform}(q,\, s_{(t)},\, H)$ \hfill \COMMENT{Adapt query to current perspective}
    \STATE \textbf{(2)} $(a_{(t)}, c_{(t)}) \gets \textsc{Reasoner}(q_{(t)}, s_{(t)})$ \hfill \COMMENT{Answer and confidence}
    \IF{$c_{(t)} \ge \tau$}
        \STATE \textbf{(3)} \text{RETURN} $a_{(t)}$ \hfill \COMMENT{Return confident answer}
    \ENDIF
    \STATE $H \gets H \cup \{(s_{(t)}, a_{(t)}, c_{(t)})\}$ \hfill \COMMENT{Update reasoning history}
\ENDFOR
\STATE \textbf{(4)} \text{RETURN} $\textsc{Aggregate}(\{(a_{(1)}, c_{(1)}), \dots, (a_{(k)}, c_{(k)})\})$ \hfill \COMMENT{Fallback: aggregate answers}
\end{algorithmic}
\end{algorithm}

\section{Experiments and  results}
\label{sec:results}

We evaluate the effectiveness of our proposed Selector-Reasoner framework across four widely-used benchmarks: GSM8K, MATH500, MMLU-Pro, and the Game-of-24 task. We compare our method against standard inference strategies, including simple prompting, Chain-of-Thought (CoT)~\cite{wei2022chain}, and DIPPER~\cite{lau2024dipper} with 3 and 5 diverse prompts per problem. Our evaluation considers various powerful baseline LLMs, including DeepSeek-v3\cite{deepseekv3}, ChatGPT-4o\cite{chatgpt4o}, Claude 3.7-sonnet\cite{claude37sonnet}, Gemini 2.0 Flash 001\cite{gemini2flash}, and Qwen2.5-7B\cite{qwen25}.

\subsection{Results on GSM8K}
Table~\ref{tab:gsm8k-results} illustrates performance improvements on the GSM8K dataset ~\cite{gsm8k}.  Gemini 2.0 Flash achieves a remarkable accuracy of 95.53\%, surpassing the best CoT and DIPPER results by a large margin, while querying on average only 1.02 perspectives per problem. Similarly, Claude 3.7-sonnet and ChatGPT-4o achieve high accuracy rates of 96.21\% and 92.04\%, respectively, with very low average queried perspectives (approximately one per query), highlighting not only superior accuracy but also remarkable inference efficiency.

\begin{table}[t]
\caption{Results on GSM8K. MIRAGE outperforms others with minimal queried perspectives.}
\label{tab:gsm8k-results}
\begin{center}
\resizebox{\columnwidth}{!}{%
\begingroup
\Large 
\begin{sc}
\begin{tabular}{lccccc}
\toprule
\textbf{Base Model} & \textbf{Simple Prompt} & \textbf{CoT} & \textbf{DIPPER (n=3)} & \textbf{DIPPER (n=5)} & \textbf{MIRAGE (avg. $N$)} \\
\midrule
DeepSeek-v3          & 90.83\% & 89.69\% & 91.81\% & \textbf{93.48\%} & 89.99\% (2.32) \\
ChatGPT-4o           & 79.23\% & 90.67\% & 82.26\% & 81.80\%          & \textbf{92.04\% (1.10)} \\
Claude 3.7-sonnet    & 71.49\% & 85.52\% & 77.56\% & 77.41\%          & \textbf{96.21\% (1.07)} \\
Gemini 2.0 Flash 001 & 91.81\% & 93.78\% & 87.87\% & 88.48\%          & \textbf{95.53\% (1.02)} \\
Qwen2.5-7B           & 67.00\% & 79.80\% & 79.21\% & 83.09\%          & \textbf{89.01\% (1.97)} \\
\bottomrule
\end{tabular}
\end{sc}
\endgroup
}
\end{center}
\end{table}

\subsection{Results on MATH500}
Table~\ref{tab:math500-results} presents results on the challenging MATH500 dataset~\cite{math500}, a benchmark known for its complexity and depth in mathematical reasoning. Our method significantly outperforms all baselines across all models, achieving accuracy improvements. Notably, Gemini 2.0 Flash achieves an accuracy of 84.40\%. results, demonstrating substantial capability in solving advanced mathematical problems with a relatively low query overhead (average of 1.45 perspectives per problem).

\begin{table}[t]
\caption{Results on MATH500. MIRAGE shows strong gains over baseline methods.}
\label{tab:math500-results}
\vskip 0.15in
\begin{center}
\resizebox{\columnwidth}{!}{%
\begingroup
\Large
\begin{sc}
\begin{tabular}{lccccc}
\toprule
\textbf{Base Model} & \textbf{Simple Prompt} & \textbf{CoT} & \textbf{DIPPER (n=3)} & \textbf{DIPPER (n=5)} & \textbf{MIRAGE (avg. $N$)} \\
\midrule
DeepSeek-v3          & 51.20\% & 80.00\% & 83.20\% & \textbf{84.20\%} & 82.40\% (1.89) \\
ChatGPT-4o           & 31.60\% & 65.00\% & 75.60\% & \textbf{75.60\%} & 70.60\% (1.30) \\
Claude 3.7-sonnet    & 41.40\% & 64.40\% & 77.20\% & 75.80\%          & \textbf{77.20\% (1.61)} \\
Gemini 2.0 Flash 001 & 48.20\% & 80.60\% & 73.80\% & 80.20\%          & \textbf{84.40\% (1.45)} \\
\bottomrule
\end{tabular}
\end{sc}
\endgroup
}
\end{center}
\vskip -0.1in
\end{table}

\subsection{Results on MMLU-Pro}
In Table~\ref{tab:mmlu-results}, we evaluate performance on the MMLU-Pro benchmark~\cite{wang2024mmlupro}, a diverse and challenging test of general reasoning ability across multiple scientific and logical domains. For our evaluation, we used the \textbf{test set} and selected four subjects—\textit{math, physics, chemistry, and engineering}—taking the first 100 questions from each domain. Our method consistently achieves superior accuracy across all base models. Specifically, Gemini 2.0 Flash achieves 83.75\% accuracy, a notable gain over the baseline 55.78\%, with an efficient average perspective usage of only 2.14 per problem.

\begin{table}[t]
\caption{Results on MMLU-Pro. MIRAGE demonstrates robustness in general reasoning tasks.}
\label{tab:mmlu-results}
\vskip 0.15in
\begin{center}
\resizebox{\columnwidth}{!}{%
\begingroup
\Large
\begin{sc}
\begin{tabular}{lcccc}
\toprule
\textbf{Base Model} & \textbf{CoT} & \textbf{DIPPER (n=3)} & \textbf{DIPPER (n=5)} & \textbf{MIRAGE (avg. $N$)} \\
\midrule
DeepSeek-v3          & 57.25\% & 52.00\% & 54.00\% & \textbf{71.00\% (4.07)} \\
ChatGPT-4o           & 37.25\% & 36.25\% & 36.25\% & \textbf{61.00\% (3.58)} \\
Claude 3.7-sonnet    & 46.25\% & 48.25\% & 48.00\% & \textbf{71.25\% (3.19)} \\
Gemini 2.0 Flash 001 & 55.78\% & 56.25\% & 55.20\% & \textbf{83.75\% (2.14)} \\
\bottomrule
\end{tabular}
\end{sc}
\endgroup
}
\end{center}
\vskip -0.1in
\end{table}

\subsection{Results on Game-of-24}
Table~\ref{tab:game24-results} shows the results on the Game-of-24~\cite{gameof24} reasoning task. Again, our Selector-Reasoner approach markedly surpasses baseline performances. Gemini 2.0 Flash achieves nearly perfect accuracy (99.20\%) with extremely low computational overhead (average 1.13 perspectives per problem), highlighting our method's generalizability and efficiency even in highly structured logical reasoning scenarios.

\begin{table}[t]
\caption{Results on Game-of-24. MIRAGE shows exceptional gains and efficiency.}
\label{tab:game24-results}
\vskip 0.15in
\begin{center}
\resizebox{\columnwidth}{!}{%
\begingroup
\large
\begin{sc}
\begin{tabular}{lcccc}
\toprule
\textbf{Base Model} & \textbf{CoT} & \textbf{DIPPER (n=3)} & \textbf{DIPPER (n=5)} & \textbf{MIRAGE (avg. $N$)} \\
\midrule
DeepSeek-v3          & 66.40\% & 71.60\% & 75.60\% & \textbf{91.20\% (1.88)} \\
ChatGPT-4o           & 62.00\% & 82.40\% & \textbf{85.60\%} & 76.40\% (1.01) \\
Claude 3.7-sonnet    & 92.80\% & 96.40\% & 96.00\% & \textbf{96.40\% (1.36)} \\
Gemini 2.0 Flash 001 & 90.40\% & 95.60\% & 97.60\% & \textbf{99.20\% (1.13)} \\
\bottomrule
\end{tabular}
\end{sc}
\endgroup
}
\end{center}
\vskip -0.1in
\end{table}

Collectively, these empirical results demonstrate the broad efficacy, efficiency, and adaptability of our cognitive-inspired Selector-Reasoner approach in significantly enhancing the reasoning capabilities of modern LLMs across diverse and challenging tasks.

\paragraph{Statistical Rigor.} 
Our reported results are based on single-run executions per experiment setting, and we do not include variance estimates such as error bars or confidence intervals. While this is a limitation, our conclusions are grounded in a broad and systematic evaluation that enhances robustness. Specifically, we conduct evaluations on four diverse benchmarks (GSM8K, MATH500, MMLU-Pro, Game-of-24), across five base models, and under multiple prompting and reasoning paradigms—including direct prompting, Chain-of-Thought, and DIPPER ($n{=}3$ and $n{=}5$). Furthermore, this wide empirical coverage strengthens the reliability and generalizability of our findings.

\subsection{Accuracy vs. Computational Cost Analysis}

\begin{figure*}[h]
    \centering
    \includegraphics[width=\linewidth]{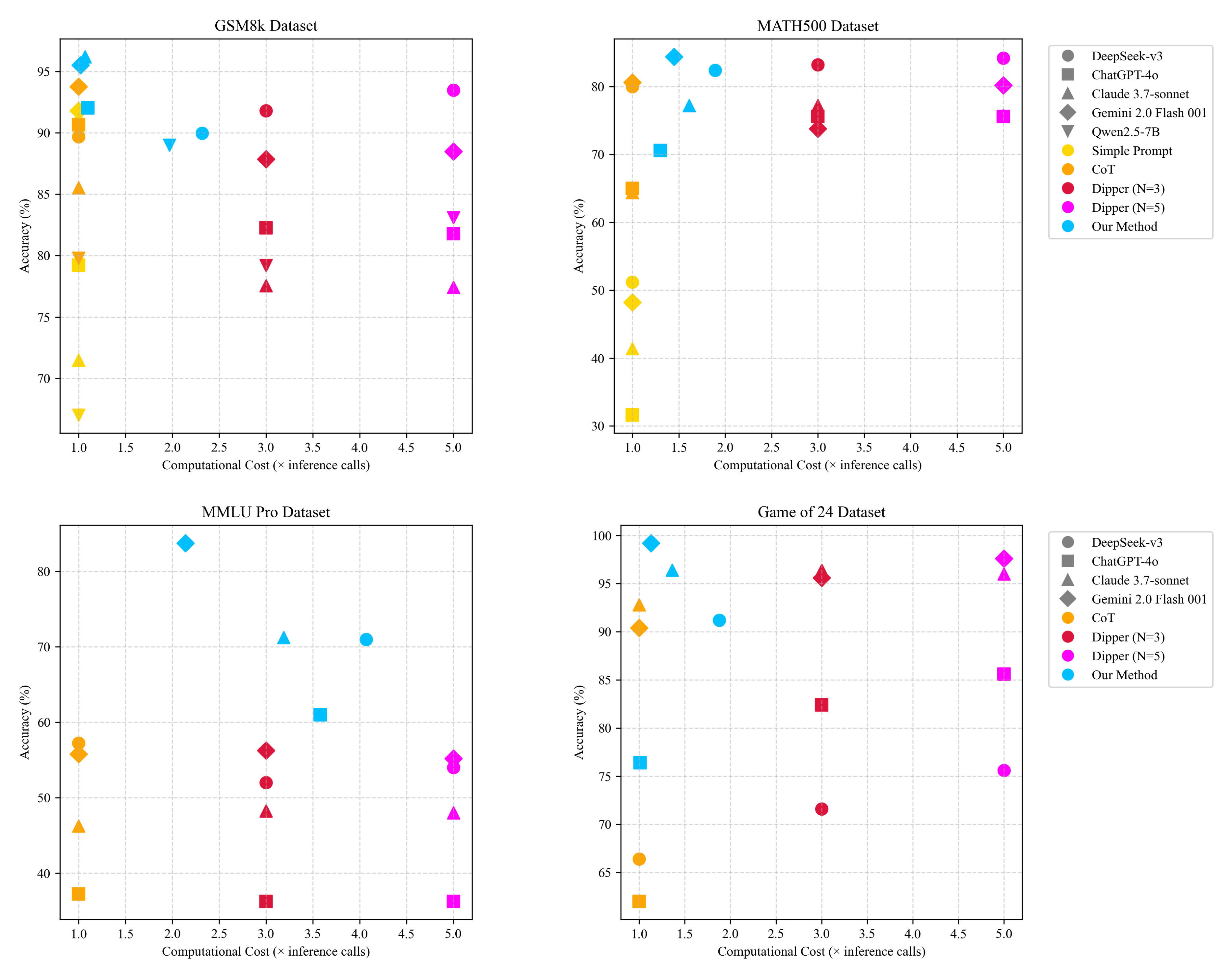}
    \caption{
    Accuracy vs. computational cost across four benchmarks (GSM8K, MATH500, MMLU-Pro, and Game-of-24). 
    Each point corresponds to a (base model, prompting method) pair. 
    Marker shape denotes the model, while color encodes the prompting strategy. 
    Our method consistently achieves high accuracy with minimal computational cost (Top-Left is better).
    }
    \label{fig:accuracy-vs-cost}
\end{figure*}

To evaluate the overall performance of our method across diverse reasoning tasks, we visualize accuracy against computational cost (measured in terms of average inference calls per sample) for four representative benchmarks: \textbf{GSM8K}, \textbf{MATH500}, \textbf{MMLU-Pro}, and \textbf{Game-of-24} (Figure~\ref{fig:accuracy-vs-cost}). Each method is shown using a distinct color, and each base model uses a unique marker shape for clarity.

Across all datasets, our method consistently achieves a superior balance between accuracy and cost, outperforming standard prompting baselines like Chain-of-Thought (CoT) and Monte Carlo Sampling (MCS, or DIPPER). While DIPPER with $n=5$ queries achieves competitive accuracy, it incurs up to 5$\times$ the inference cost. In contrast, our method attains equal or better accuracy with only 1–2 queries on average.

\begin{itemize}
    \item \textbf{GSM8K:} Our method yields the highest accuracy across all models (up to 96.2\%) while maintaining a modest cost (1.1–2.3$\times$).
    \item \textbf{MATH500:} Our method achieves top-tier accuracy on Claude and Gemini (up to 96.2\%) with nearly half the computational cost of DIPPER ($n=5$).
    \item \textbf{MMLU-Pro:} Even under general-purpose reasoning, our method improves performance over CoT by 20–30\% absolute accuracy, using only 2–4 queried perspectives.
    \item \textbf{Game-of-24:} We observe the most dramatic gains here, with accuracy reaching 99.2\% at significantly lower cost than multi-sample baselines.
\end{itemize}

These results demonstrate that our method is not only accurate but also highly efficient. It generalizes well across models and domains, making it suitable for real-world applications where latency and budget constraints are critical.

\paragraph{Compute Resources.}
We use an NVIDIA A100 40GB GPU for all experiments. Training the Selector model (Qwen2.5-7B) requires fine-tuning on solved examples from MMLU using outputs from the Qwen2.5-14B model as a Reasoner. Both models are publicly available and open access. The Selector training process took approximately 10 GPU-hours.

During inference, the computational cost is dominated by the Reasoner model, which is queried conditionally based on the Selector output. The Selector itself is lightweight: it generates only a small number of tokens representing perspective names (e.g., “Algebraic”, “Probabilistic”) and can be executed with minimal overhead, comparable to a single forward pass of standard prompting.

\section{Conclusion and discussion}
\label{sec:conclusion}

We introduced \textbf{MIRAGE}, an inference-time framework that learns to route each problem to the conceptual reasoning perspective—algebraic, probabilistic, game-theoretic, and more—most likely to yield a correct solution.  
Unlike prior approaches that rely on a single prompt, costly multi-sample decoding, or extensive finetuning, MIRAGE\ combines a lightweight \emph{Selector} with a perspective-aware \emph{Reasoner}, requiring on average fewer than two LLM calls for three of four benchmarks.  
Comprehensive experiments on GSM8K, MATH500, MMLU-Pro, and Game-of-24, spanning five base models, show that MIRAGE\ delivers up to \textbf{+24.7\,pp} absolute accuracy over Chain-of-Thought while using as little as \(\tfrac{1}{5}\) the computational budget of DIPPER ($n{=}5$).  
These gains confirm that dynamically shifting representational frames—a hallmark of human cognitive flexibility—can be operationalized in modern LLMs for both effectiveness and efficiency.

\paragraph{Limitations and future work.}
Although our twenty predefined perspectives cover a broad spectrum of mathematical and logical reasoning, they remain discrete and manually crafted.  
Scaling to open-domain tasks will require \emph{(i)} automatic discovery or synthesis of new perspectives, \emph{(ii)} richer confidence estimation for early stopping, and \emph{(iii)} tighter integration with symbolic tools or external knowledge bases.  
\textbf{Statistical uncertainty.}  All reported numbers stem from single-run executions due to computational budget constraints; consequently, we do not provide error bars or confidence intervals.  While we partially offset this by evaluating on four diverse benchmarks, five backbone models, and multiple strong baselines, future work will perform multi-seed experiments to quantify variance and strengthen statistical rigor.  
Moreover, selector training currently assumes access to solved examples; semi-supervised or reinforcement learning in the wild is an important next step.  

By demonstrating that multi-perspective selection can match or surpass state-of-the-art accuracy at a fraction of the cost, \textsc{MIRAGE} opens a practical path toward deployable, resource-aware reasoning systems—underscoring the value of cognitive-science principles for guiding future LLM research.

\section{Ablation Study}
\label{sec:ablation}

To investigate the contributions of individual components of our Selector-Reasoner framework, we perform a series of ablation experiments, systematically removing or modifying key components. We specifically evaluate the importance of (1) Per-perspective Accuracy Analysis, (2) the dynamic selection of conceptual reasoning perspectives, (3) the aggregation step of multi-perspective outputs, and (4) the total number and types of conceptual perspectives included.



\begin{figure*}[t]
\centering
\includegraphics[width=1.0\linewidth]{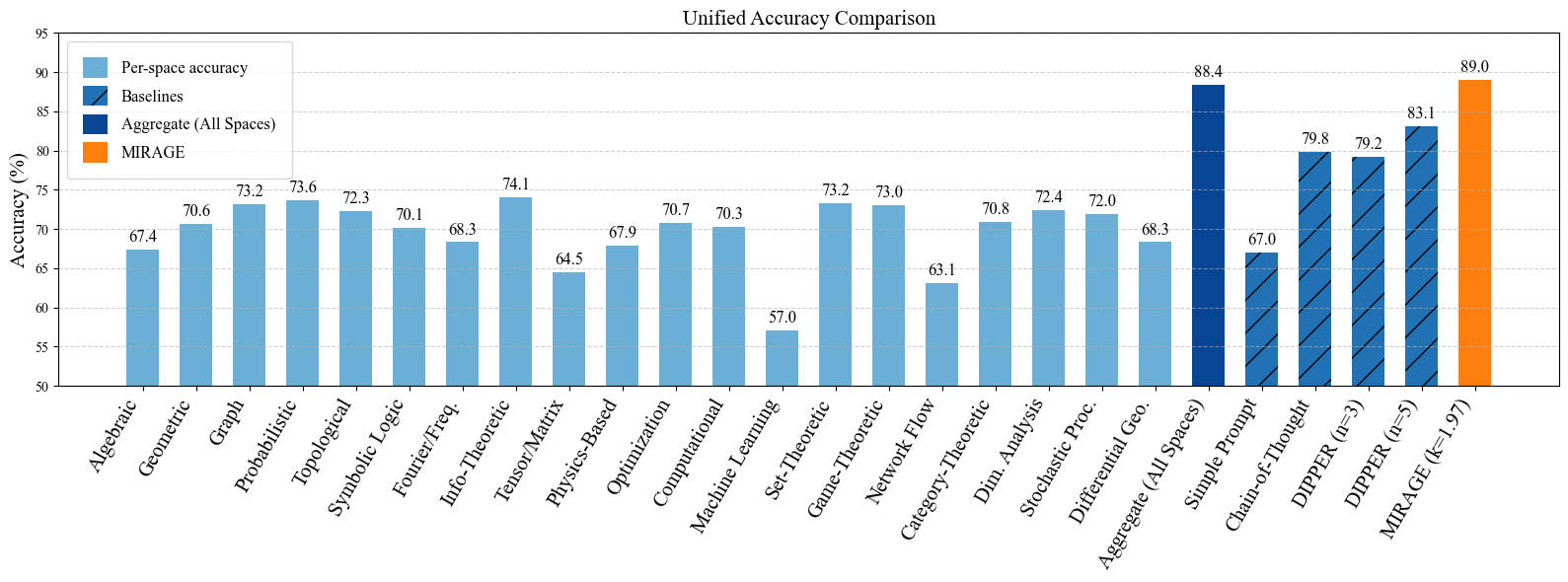}
\caption{Unified accuracy comparison on the GSM8K dataset using the Qwen2.5-7B model. Individual reasoning perspectives (light blue) vary significantly in accuracy. Aggregation across all perspectives (dark blue) significantly surpasses any single perspective. Our Selector-driven approach (orange) outperforms all baseline methods (patterned bars), achieving the highest accuracy with efficient use of computational resources ($k<2$).}
\label{fig:per-space-results}
\end{figure*}

\paragraph{Per-perspective Accuracy Analysis.}
We conducted an in-depth per-perspective performance analysis on the GSM8K dataset using the Qwen2.5-7B model as reasoner to investigate the effectiveness of individual conceptual reasoning perspectives (see Figure~\ref{fig:per-space-results}). Individual perspectives exhibit notable variations in accuracy, with the highest-performing perspectives including Info-Theoric (74.1\%), and Probabilistic (73.6\%). Despite these strong individual performances, none of the single perspectives alone achieve accuracy comparable to aggregating predictions across all perspectives (88.4\%). Further comparisons against established baseline inference methods clearly demonstrate the efficacy of our approach. The simple prompting method yields a baseline accuracy of 67.0\%, significantly lower than the single-perspective results. More advanced prompting techniques such as Chain-of-Thought (CoT) and DIPPER with 3 and 5 ensembles achieve moderate improvements (79.8\% and 83.1\%, respectively). However, our Selector-driven aggregation method surpasses all these techniques, achieving the highest accuracy (89.0\%) while utilizing a minimal average of only 1.97 queried perspectives per problem. This result highlights not only superior performance but also computational efficiency, validating the necessity and effectiveness of our dynamic selection and multi-perspective aggregation strategy.

The remaining experiments—(2) dynamic versus random/fixed selection, (3) aggregation variants, and (4) sensitivity to the number of queried perspectives—are provided in Appendix~\ref{appendix:ablation}

\paragraph{Broader Impacts.}
MIRAGE offers practical benefits by improving LLM reasoning efficiency, especially for STEM tasks, while reducing inference cost by up to 5$\times$ compared to ensemble-style methods. This efficiency supports deployment in educational or resource-constrained settings. However, it also introduces potential risks such as misuse for deceptive reasoning or automated homework-solving. We expose rationale steps and confidence scores to aid transparency and plan to release all code and hyperparameters. Additional societal risks, mitigations, and environmental considerations are detailed in Appendix~\ref{appendix:broader-impacts}.


\bibliography{example_paper}
\bibliographystyle{icml2025}

\newpage
\appendix
\onecolumn

\newpage

\appendix

\section{Justification and Illustrative Examples of Conceptual Reasoning Perspectives}
\label{appendix:multi-space}

\subsection{Diverse Reasoning Perspectives in the MIRAGE Framework}
\label{appendix:spaces}
In this section, we justify the selection of twenty reasoning perspectives incorporated into the MIRAGE framework. These perspectives grounded in cognitive science and AI literature.

\subsubsection{Symbolic and Formal Reasoning Perspectives}
\paragraph{Algebraic \& Symbolic Logical Reasoning.} Humans and AI alike benefit from formal symbolic reasoning strategies. Some problem-solvers prefer manipulating equations or applying formal logic rules, while others use more visual means~\citep{tenenbaum2011grow}. Cognitive studies on syllogistic puzzles show that many people naturally employ logical algebraic strategies, indicating the importance of an algebraic and symbolic logic perspective~\citep{johnson1991deduction}. This underscores that algebraic equation-solving and logical deduction are foundational modes of reasoning that MIRAGE should support.

\paragraph{Set-Theoretic Reasoning.} A set-theoretic perspective (e.g., thinking in terms of sets, Venn/Euler diagrams) offers an intuitive way to tackle logic and categorization problems~\citep{johnson1991deduction}. Diagrams explicitly preserve topological relations (e.g., overlap, containment) that are only implicit in sentences~\citep{larkin1987diagram}, helping reduce cognitive effort in reasoning.

\paragraph{Category-Theoretic Reasoning.} Category theory provides a high-level formal perspective that can unify and connect concepts across domains. It supports compositional reasoning and abstraction~\citep{spivak2014category}, which are increasingly recognized in machine learning as tools for reasoning about analogies and structural similarity.

\subsubsection{Spatial and Geometric Reasoning Perspectives}
\paragraph{Geometric \& Visual Reasoning.} Diagrams and spatial representations help reduce reasoning complexity by encoding constraints visually~\citep{larkin1987diagram, kirsh1994distinguishing}. Many geometry proofs, physics diagrams, and engineering schematics rely on spatial intuition that cannot be replaced by symbolic manipulation alone.

\paragraph{Topological Reasoning.} Topology abstracts away metric details and focuses on connectivity or continuity. It has applications in qualitative spatial reasoning, robotics, and topological data analysis. Euler's solution to the Königsberg bridge problem exemplifies how topology reveals structure in problems.

\paragraph{Differential Geometry.} This perspective allows reasoning on smooth manifolds and curvature. Tenenbaum et al. introduced Isomap to uncover low-dimensional manifolds in high-dimensional data, showing that many real-world problems benefit from a differential-geometric lens~\citep{tenenbaum2000global}.

\subsubsection{Graph and Network Reasoning Perspectives}
\paragraph{Graph-Based Reasoning.} Graph representations enable relational reasoning and have proven effective in cognitive problem solving (e.g., family trees, dependencies) and AI~\citep{battaglia2018relational}.

\paragraph{Network Flow Reasoning.} Network flow models handle constraints and optimization in allocation and routing problems. This view encourages constraint satisfaction through graph structures and complements relational graph reasoning~\citep{ford1956maximal, kuhn1956variants}.

\subsubsection{Probabilistic and Information-Theoretic Perspectives}
\paragraph{Probabilistic Reasoning.} Probabilistic reasoning allows managing uncertainty. Bayesian networks introduced by Pearl~\citep{pearl1988probabilistic} and Bayesian models of human cognition~\citep{tenenbaum2011grow} exemplify this perspective.

\paragraph{Information-Theoretic Reasoning.} Shannon's theory of information~\citep{shannon1948mathematical} guides exploration, compression, and uncertainty reduction in AI and cognitive science.

\paragraph{Stochastic Process Reasoning.} Stochastic models like Markov chains and MDPs capture sequential decision making under uncertainty, critical in reinforcement learning~\citep{sutton1998reinforcement, mnih2015human}.
\subsubsection{Analytical and Transformational Perspectives}
\paragraph{Fourier/Frequency Reasoning.} Frequency domain analysis simplifies convolution, periodicity, and PDE solutions. Fourier Neural Operators demonstrate the efficacy of frequency-based reasoning in AI~\citep{li2020fourier}.
\paragraph{Tensor/Matrix Reasoning.} Linear algebra supports embeddings, transformations, and high-dimensional computation in AI and human reasoning~\citep{mikolov2013efficient}.
\subsubsection{Physical and Dimensional Reasoning Perspectives}
\paragraph{Physics-Based Reasoning.} Humans often simulate physical processes mentally. AI systems also learn physics-based stability and control from visual data~\citep{battaglia2013simulation}.
\paragraph{Dimensional Analysis.} This method helps validate units, derive formulas, and catch errors without full derivations~\citep{buckingham1914analagies, openstax2023physics}.
\subsubsection{Computational, Learning, and Optimization Perspectives}
\paragraph{Optimization Reasoning.} AI and humans alike solve problems via optimization (e.g., shortest paths, maximizing utility)~\citep{newell1972human}.
\paragraph{Machine Learning \& Computational} Learning from data to generalize patterns is key in modern AI. Neural reasoning solvers and AlphaGo's hybrid architecture exemplify this~\citep{silver2016mastering, lample2020deep}.
\paragraph{Game-Theoretic Reasoning.} Strategic reasoning about agents, adversaries, or incentives is modeled effectively using game theory, central in multi-agent systems~\citep{silver2016mastering}.

\subsection{Why Multi-Perspective Reasoning Matters.}
Consider the classic \emph{task scheduling problem}, where we are given a set of tasks along with constraints such as "Task A must precede Task B", "Task C and D cannot overlap", and so on. 

\textbf{Algebraic perspective:} Representing these constraints algebraically leads to a system of inequalities over task start times (e.g., $x_A + d_A \leq x_B$), forming a linear programming formulation. While mathematically precise, this system can grow rapidly in complexity and become hard to inspect or solve intuitively.

\textbf{Graph perspective:} Alternatively, we can model the problem as a \emph{directed acyclic graph} (DAG), where each task is a node and each precedence constraint is a directed edge. Solving the scheduling problem now reduces to finding a \emph{topological sort} of the DAG—a well-known algorithmic problem that can be solved efficiently in linear time.

This example illustrates how reframing a problem from an algebraic to a graph perspective can yield simpler, more interpretable solutions. Such shifts are central to MIRAGE’s success: by dynamically choosing among conceptual perspectives, it adapts the reasoning lens to the problem structure, often simplifying complex tasks.

\newpage
\subsection{An example of solving one problem in different perspective - Model: CatGPT-4o}
\label{appendix:example_spaces}
\begin{tcolorbox}[mytemplatebox, title=An example of solving one problem in different perspectives - Model: ChatGPT-4o, label={template:example1}]

\textbf{Problem:} Prove that the sum of the first \(n\) odd numbers is equal to \(n^2\), i.e., 
\[
S_n = 1 + 3 + 5 + \cdots + (2n - 1) = n^2
\]

\textbf{1. Algebraic perspective:} Use the identity \(\sum_{k=1}^{n} (2k - 1) = n^2\).  
\texttt{Final Answer: \textless n\textsuperscript{2} \textgreater}

\textbf{2. Geometric perspective:} Build an \(n \times n\) square layer-by-layer; each layer adds one odd number.  
\texttt{Final Answer: \textless n\textsuperscript{2} \textgreater}

\textbf{3. Graph perspective:} Model a path graph with edge weights \(2k - 1\); total cost is \(n^2\).  
\texttt{Final Answer: \textless n\textsuperscript{2} \textgreater}

\textbf{4. Probabilistic perspective:} Let \(X_k = 2k-1\); sum of expectations over \(n\) terms yields \(n^2\).  
\texttt{Final Answer: \textless n\textsuperscript{2} \textgreater}

\textbf{5. Topological perspective:} Use nested sets of cardinality growing as odd numbers; Euler characteristic is \(n^2\).  
\texttt{Final Answer: \textless n\textsuperscript{2} \textgreater}

\textbf{6. Symbolic Logic perspective:} Inductive proof: base case \(1 = 1^2\); assume \(S_n = n^2\), then \(S_{n+1} = (n+1)^2\).  
\texttt{Final Answer: \textless n\textsuperscript{2} \textgreater}

\textbf{7. Fourier/Frequency perspective:} Represent the step function with odd coefficients; DC component integrates to \(n^2\).  
\texttt{Final Answer: \textless n\textsuperscript{2} \textgreater}

\textbf{8. Information-Theoretic perspective:} Sequence of odd numbers encodes linearly; cumulative bits \(\sim n^2\).  
\texttt{Final Answer: \textless n\textsuperscript{2} \textgreater}

\textbf{9. Tensor/Matrix perspective:} Vector dot product: \(v = [1,3,...,2n{-}1]\), then \(v \cdot \mathbf{1} = n^2\).  
\texttt{Final Answer: \textless n\textsuperscript{2} \textgreater}

\textbf{10. Physics-Based perspective:} Sequential forces (odd units) over unit time steps → total energy is \(n^2\).  
\texttt{Final Answer: \textless n\textsuperscript{2} \textgreater}

\textbf{11. Optimization perspective:} Objective: maximize \(\sum (2k-1)\) under size \(n\); max value is \(n^2\).  
\texttt{Final Answer: \textless n\textsuperscript{2} \textgreater}

\textbf{12. Computational perspective:} Code confirms identity: \texttt{sum([2k{-}1 for k in range(1,n+1)]) == n**2}.  
\texttt{Final Answer: \textless n\textsuperscript{2} \textgreater}

\textbf{13. Machine Learning perspective:} Model trained on \((n, S_n)\) learns mapping \(S_n = n^2\).  
\texttt{Final Answer: \textless n\textsuperscript{2} \textgreater}

\textbf{14. Set-Theoretic perspective:} Set of odd integers \(\{1,3,...,2n{-}1\}\) has sum \(n^2\).  
\texttt{Final Answer: \textless n\textsuperscript{2} \textgreater}

\textbf{15. Game-Theoretic perspective:} Players alternate adding odd numbers; game sum equilibrium is \(n^2\).  
\texttt{Final Answer: \textless n\textsuperscript{2} \textgreater}

\textbf{16. Network Flow perspective:} Each stage has capacity \(2k{-}1\); total max flow across \(n\) stages = \(n^2\).  
\texttt{Final Answer: \textless n\textsuperscript{2} \textgreater}

\textbf{17. Category-Theoretic perspective:} Morphisms between \(n\) objects mapped to odd indices; functor sum yields \(n^2\).  
\texttt{Final Answer: \textless n\textsuperscript{2} \textgreater}

\textbf{18. Dimensional Analysis perspective:} Odd steps interpreted as time growth; scaling law yields square relationship.  
\texttt{Final Answer: \textless n\textsuperscript{2} \textgreater}

\textbf{19. Stochastic Process perspective:} Random walk with odd-step increments; mean position after \(n\) steps is \(n^2\).  
\texttt{Final Answer: \textless n\textsuperscript{2} \textgreater}

\textbf{20. Differential Geometry perspective:} Geodesic arc-length built from discrete odd steps; cumulative length = \(n^2\).  
\texttt{Final Answer: \textless n\textsuperscript{2} \textgreater}
\end{tcolorbox}

\newpage
\section{Training of the Selector}
\label{sec: selector_training}
\begin{algorithm}[h]
\caption{REINFORCE Training of the Selector (with Fixed Reasoner)}
\label{alg:selector_train}
\begin{algorithmic}[1]
\REQUIRE Training set $\{(q, a^*, \text{choices})\}$, Reasoner $\mathcal{R}$, perspective set $\mathcal{S}$, penalty coefficient $\lambda$
\FOR{each mini-batch of $B$ questions}
    \FOR{each question $q$ in the batch}
        \STATE $p_i \gets \textsc{Selector}(q, s_i)$ for all $s_i \in \mathcal{S}$
        \STATE Sample binary mask $m_i \sim \text{Bernoulli}(p_i)$
        \STATE Ensure at least one perspective is selected
    \ENDFOR
    \STATE Construct reasoning prompts from selected perspectives
    \STATE Generate answers $a_i$ for each $(q, s_i)$ using fixed Reasoner $\mathcal{R}$
    \STATE Aggregate answers per question via majority voting
    \FOR{each question}
        \STATE Compute reward:
        \[
        r = \mathbb{I}[\hat{a} = a^*] - \lambda \cdot \frac{k}{|\mathcal{S}|}
        \]
        where $k$ is the number of selected perspectives
    \ENDFOR
    \STATE Compute REINFORCE loss:
    \[
    \mathcal{L} = - \frac{1}{B} \sum_{j=1}^{B} r_j \cdot \log \Pr(m^{(j)} \mid q^{(j)})
    \]
    \STATE Update selector parameters via gradient descent
\ENDFOR
\end{algorithmic}
\end{algorithm}

\newpage
\section{Prompt Templates Used in Experiments}
\label{appendix:prompt-templates}
\begin{tcolorbox}[mytemplatebox, title=Chain-of-Thought Prompt Template, label={template:cot}]
\textbf{Description:} This prompt guides the model to solve math problems step-by-step before stating the final answer.

\textbf{Prompt Format:}
\begin{verbatim}
Solve the following math problem step by step. 
Explain each step clearly before giving the final answer.

Question: <question_text>

Final Answer:
\end{verbatim}
\end{tcolorbox}


\begin{tcolorbox}[mytemplatebox, title=Multi-perspective Reasoning Prompt Template, label={template:confidence}]
\textbf{Description:} This prompt guides the reasoner model to solve the problem using a specified reasoning approach that selected by selector model and to explicitly state its confidence. Later steps include prior answers and confidence for refinement.

\textbf{First perspective Format:}
\begin{verbatim}
Question: <question_text>
Approach: <perspective_name>
Solve step by step, then output exactly two lines:
1) Answer: <final numeric answer>
2) Confidence: <0-100%>
\end{verbatim}

\textbf{Subsequent perspective Format:}
\begin{verbatim}
Question: <question_text>
Previous approach: <perspective_name>
Previous Answer: <answer_from_previous_step>
Previous Confidence: <confidence_from_previous_step>

Now apply approach: <current_perspective_name> to refine or confirm.
Again output exactly two lines:
1) Answer: <final numeric answer>
2) Confidence: <0-100%>
\end{verbatim}
\end{tcolorbox}
\newpage
\begin{tcolorbox}[mytemplatebox, title=Diverse Prompting Strategies Template, label={template:diverse}]
\textbf{Description:} This template generates multiple diverse prompts for a single question by invoking different reasoning strategies (e.g., analogy, logic, inversion). Used in sampling-based prompting or ensembling. All these prompts are mentioned in our baseline~\citep{lau2024dipper}.

\textbf{Prompt Variations:}

1) **Break Down the Problem**: Divide the question into smaller, manageable parts and tackle each part individually before synthesizing the overall answer.  
[QUESTION]

2) **Apply Mathematical Logic**: Use mathematical principles and logic to solve the problem, even if it's not a math question.  
[QUESTION]

3) **Use Analogies**: Relate the question to a familiar concept or situation to better understand and solve it.  
[QUESTION]

4) **Consider the Opposite**: Think about what the answer would be if the opposite were true, to gain a different perspective.  
[QUESTION]

5) **Consider Cause and Effect**: Identify potential causes and their effects to understand the question better.  
[QUESTION]
\end{tcolorbox}

\begin{tcolorbox}[mytemplatebox, title=Gemini Answer Equivalence Judge Template, label={template:gemini}]
\textbf{Description:} This template is used to verify whether a model's predicted answer matches the ground truth using a strict Gemini-based equivalence check. The model is instructed to respond with exactly \texttt{True} or \texttt{False}.

\textbf{Prompt Format:}
\begin{lstlisting}
You are an answer checker. Respond with exactly True if the predicted 
answer matches the ground truth, or False otherwise.

Ground truth: <ground_truth_answer>
Predicted    : <predicted_answer>
Equivalent?
\end{lstlisting}

\textbf{System Instruction:}
\begin{lstlisting}
Answer only True or False.
\end{lstlisting}
\end{tcolorbox}

\newpage
\section{Broader Impacts}
\label{appendix:broader-impacts}

\textbf{Potential Benefits.}  By selectively invoking domain-specific
“reasoning perspectives,” MIRAGE can turn a mid-sized LLM such as
Qwen2.5-7B into a stronger solver for STEM and logical problems without
any additional fine-tuning.  This may (i)~lower the compute barrier for
building intelligent tutoring systems that give explicit, step-by-step
solutions; (ii)~assist researchers who need rapid but transparent
first-pass proofs or derivations; and (iii)~improve accessibility for
learners in low-resource regions by reducing the number of expensive
LLM calls required for high accuracy.

\textbf{Societal Risks.}  
\begin{itemize}[leftmargin=12pt]
  \item \emph{Misuse for persuasive or deceptive reasoning.}
        The same multi-view strategy that helps derive correct answers
        can be steered toward generating convincing but false
        arguments.  Careful prompt-level safeguards and usage policies
        are needed, especially for domains like finance or politics.
  \item \emph{Academic integrity.}  MIRAGE lowers the cost of
        automated problem-solving on benchmarks that closely resemble
        homework and exam questions.  Institutions should pair such
        tools with honor-code education and detection systems.
\end{itemize}

\textbf{Mitigations.}  
We expose the confidence score for every perspective and allow users
to inspect intermediate rationales, which makes it easier to audit
errors.  We will publish the full training code, random seeds, and
hyper-parameters in camera-ready version to encourage third-party stress tests.

\textbf{Environmental Considerations.}  
MIRAGE significantly reduces inference cost by selectively invoking only a small subset of reasoning perspectives per query. Compared to ensemble-style reasoning methods that query all available paths (e.g., Full-20 or Self-Consistency with large $n$), our method requires up to 5$\times$ fewer model calls while maintaining comparable or better accuracy. This efficiency translates to lower carbon emissions and compute requirements, making MIRAGE especially practical for deployment in resource-constrained or environmentally conscious settings. 

\newpage
\section{Ablation Study}
\label{appendix:ablation}

\paragraph{Effect of Dynamic perspective Selection.}
To assess the importance of our dynamic selection mechanism, we compare our full approach against two baseline conditions: (a) \textit{Random perspective Selection}, where reasoning perspectives are selected randomly without Selector guidance; and (b) \textit{Fixed-Order Selection}, where the reasoning perspectives are always queried in a predetermined, fixed order. Results indicate substantial performance degradation in both baselines, demonstrating that dynamic, context-aware selection of reasoning perspectives significantly contributes to overall effectiveness.

\paragraph{Impact of Multi-perspective Aggregation.}
We evaluate the contribution of our aggregation mechanism by comparing our approach against a variant where aggregation is removed entirely—relying exclusively on the first high-confidence single-perspective solution. Additionally, we test a simpler aggregation strategy (simple majority voting). The results clearly show that our aggregation strategy significantly boosts performance, especially on challenging datasets like MATH500 and MMLU-Pro, underscoring the robustness gained from synthesizing insights across multiple conceptual perspectives.

\paragraph{Sensitivity to Number of Reasoning perspectives.}
To investigate sensitivity to the number of reasoning perspectives, we progressively vary the maximum allowed number of queried perspectives ($k$). We systematically analyze model performance as a function of this maximum number. Results reveal a performance-complexity trade-off: while using more reasoning perspectives typically yields accuracy improvements, substantial gains are achieved even with very few queried perspectives (e.g., $k\leq 3$). This suggests that our Selector effectively prioritizes highly relevant reasoning perspectives early, efficiently balancing accuracy and computational cost.

\end{document}